\documentclass[10pt,journal,compsoc]{IEEEtran}

\usepackage{algorithmic}
\usepackage{array}
\usepackage{color}
\usepackage{picture}
\usepackage{subfig}
\usepackage{amsmath,amssymb}
\usepackage{mathabx}
\usepackage{graphicx}
\usepackage{epstopdf,times}

\newcommand{\argmin}{\mathop{\rm arg~min}\limits}

\ifCLASSOPTIONcompsoc
  \usepackage[nocompress]{cite}
\else
  \usepackage{cite}
\fi
\ifCLASSINFOpdf
\else
\fi
\begin{document}

\title{Mixing autoencoder with classifier: conceptual data visualization}
%
%
%
%

\author{Pitoyo Hartono
\IEEEcompsocitemizethanks{\IEEEcompsocthanksitem P. Hartono is with the Department
of Electrical and Electronics Engineering, Chukyo University, Nagoya,
Japan\protect\\
E-mail: hartono@ieee.org 
 }
\thanks{}}

\markboth{}%
{Shell \MakeLowercase{\textit{et al.}}: Bare Demo of IEEEtran.cls for Computer Society Journals}
%

\IEEEtitleabstractindextext{%
\begin{abstract}
In this short paper, a neural network that is able to form a low dimensional topological hidden representation is explained. The neural network can be trained as an autoencoder, a classifier or mix of both, and produces different low dimensional topological map for each of them. When it is trained as an autoencoder, the inherent topological structure of the data can be visualized, while when it is trained as a classifier, the topological structure is further constrained by the concept, for example the labels the data, hence the visualization is not only structural but also conceptual. The proposed neural network significantly differs from many dimensional reduction models, primarily in its ability to execute both supervised and unsupervised dimensional reduction. 

The neural network allows multi perspective visualization of the data, and thus giving more flexibility in data analysis. This paper is supported by preliminary but intuitive visualization experiments.
\end{abstract}

}

\maketitle

\IEEEdisplaynontitleabstractindextext

\IEEEpeerreviewmaketitle

\IEEEraisesectionheading{\section{Introduction}\label{sec:introduction}}

In this study, a neural network that is able to form contextual topological map in its hidden layer is explained. Over the past years rich collections of machine learning methods for visualizing high dimensional data through dimensional reduction have been proposed.  Many of them form low dimensional representation while optimizing some criteria to preserve inherent characteristics of the data. For example Stochastic Neighborhood Embedding (SNE) \cite{NIPS2002_2276} and its variants \cite{vandermaatent-sne, JMLR:v15:vandermaaten14a} reduce high dimensional data while preserving their stochastic neighborhood structure. Locally Linear Embedding (LLE) \cite{Roweis2323} is a nonliner dimensional reduction method that locally preserves the dependency of high dimensional data, while isomatric mapping (isomap) \cite{NIPS1997_1332}, \cite{Tenenbaum2319} is also a nonlinear dimensional reduction mapping that preserves the geodesic structure of high dimensional data. Kohonen's Self-organizing maps (SOM) \cite{Kohonen082,Kohonen2013} is a popular dimensional reduction and visualization method that preserves the topological structure of high dimensional data in low dimensional space. Recently, Uniform Manifold Approximation and Projection (UMAP) \cite{lel2018umap}, a manifold learning technique for dimensional reduction based on Riemannian geometry was proposed and results in high quality visualization with scalable calculation time. All those methods execute unsupervised mapping for primarily visualizing application-relevant structure of high dimensional data, but ignore the contexts (for example, labels) of the data. There are also many supervised dimensional reduction algorithms that take the context of the data into account. These methods form low dimensional representations of high dimensional data by preserving their inherent structures that are relevant to their labels. Thus, the representation is not only structural but also conceptual. Some examples of supervised dimensional reduction methods are as follows. Neighborhood Component Analysis (NCA) \cite{Goldberger:2004:NCA:2976040.2976105} that forms low dimensional representations on which the classification rate of k-nearest neighbor is maximized, a semi-supervised version of Isomap is proposed in \cite{ZHANG2018662} and a combination of Multidimensional Scaling (MDS) \cite{1671271} \cite{Kruskal1964} and SOM that can either be supervised or unsupervised was proposed in \cite{8020006}. 

While the methods above were able to generate visualization on the high dimensional data, they are either supervised or unsupervised. However, data analysis sometimes requires multi perspectives in extracting insightful information. 
Changing the methods to learn different aspect of the data is often problematic, since all the methods execute different criteria in reducing the dimension of the data. Sometimes, the dimensional reduction and the visualization are executed with different algorithms, for example in \cite{Hinton2006}, stacked-autoencoders was executed to reduce the dimension of the data, and t-SNE was executed to visualize them. It should be noted, that each dimensional reduction results in lost of information in different way, so running them in tandem will not only reduce the representation fidelity but will also infuse interpretation unclarity to the low dimensional representation space.

In this study, a neural network that is able to form low dimensional representation in its hidden layer is explained. Different from the most of the dimensional reduction method, this neural network is able to execute supervised learning, unsupervised learning or mix of them in one learning framework, by controlling a single coefficient in the learning process. The proposed neural network is built based on the previously proposed Restricted Radial Basis Function Network (r-RBF) 
\cite{hartono,hartono2016} that is a hierarchical supervised neural network that generates two dimensional topological representation in its hidden layer. Here, the output layer and the learning process is modified, so that the network can be trained as an autoencoder, a classifier or a mix of both. When the network is trained as an autoencoder, it forms a low dimensional representation that encodes a relevant topological  structure to reconstruct the high dimensional input, and thus allows the visualization on the inherent structure of the data. When it is trained as a classifier, the hidden representation is constrained by the labels of the data, so the visualization is not only structural but also conceptual, in that different labeling of the same data will produce different representation. The network can also be trained by mixing the autoencoder and classifier, resulting in flexible representations, where the difference between the inherent vectorial characteristics and the characteristics conceptualized by the labels of the data can be learned.

This short paper explains the structure and learning dynamics of the proposed neural network, and the result of the preliminary experiments.

\section{Soft-supervised Topological Autoencoder} 

\begin{figure}
\centering
\includegraphics[width =6 cm, height =6 cm]{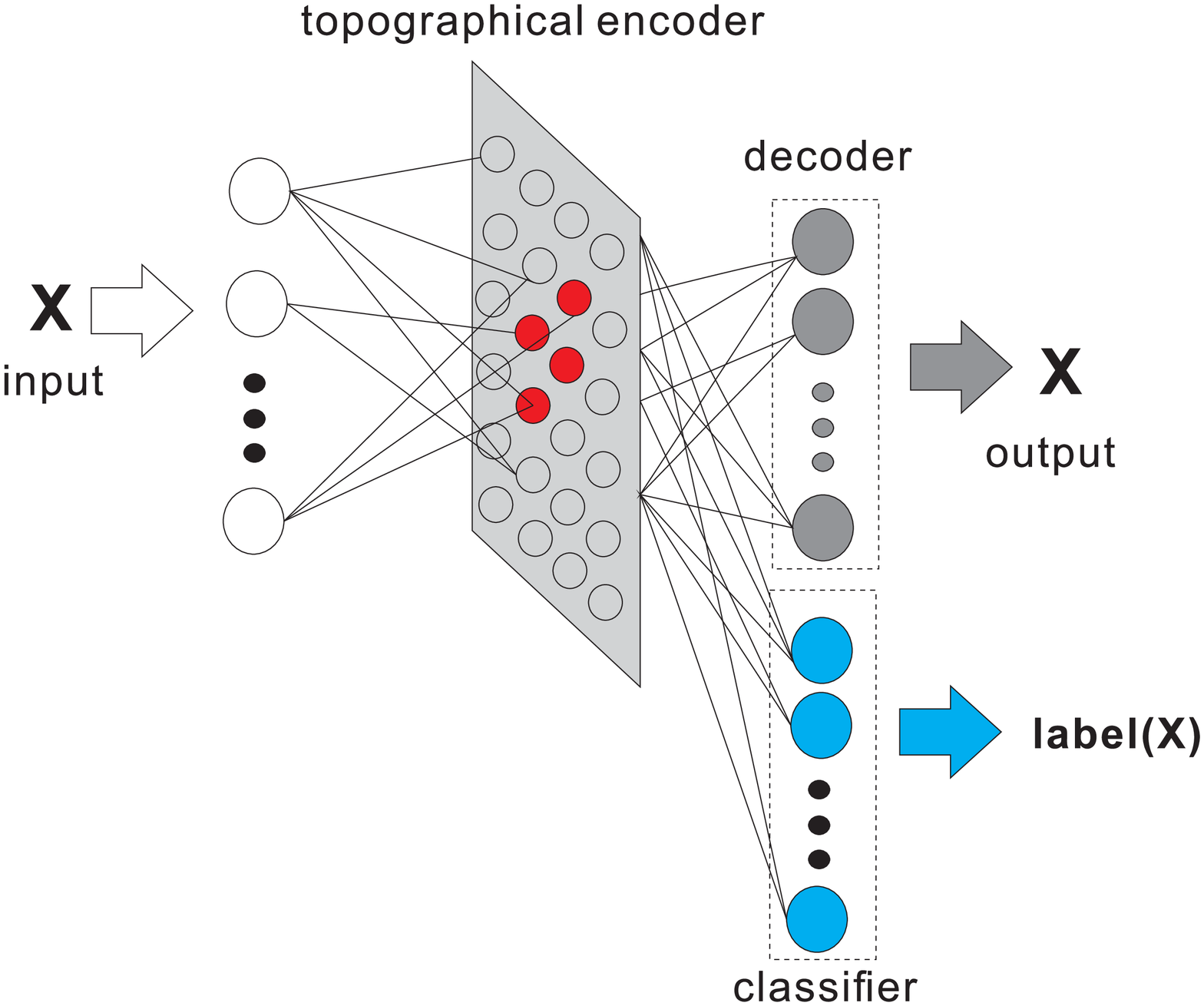}
\caption{Outline of Soft-supervised Topological Encoder}
\label{fig:outline}
\end{figure}

The outline of  Soft-supervised Topological Autoencoder (STA) is illustrated in Fig. \ref{fig:outline}. Here, a three -layered STA, in which the hidden layer is a topological layer where the neurons are allgned in two dimensional grid similar to Kohenen's Self-Organizing Maps (SOM). The output layer is composed from two parts, decoder part that reconstructs the encoded input and classifier part that predicts the labels of the input. In the training process, a mixing parameter is set to control the weightings of the decoder part and classifier part, hence the STA can be trained as an autoencoder, a classifier or a mix of both. 

Here, similar to Kohonen's SOM, due to its low dimensionality, it is possible to visualize the inputs' topographical representation, and discover the their characteristics. However, different from SOM that topologically preserves the topological structures of the high dimensional inputs into their two dimensional representation, the hidden representations are also regulated by the error signal backpropagated from the output layer. In the case that STA is trained as an autoencoder, the hidden topological representation is formed to encode topological structure that enables STA to reconstruct the inputs. In the case of a classifier, the topological structure is further constrained by the requirement to predict the output. 

The dynamics of STA is explained as follows.

\begin{eqnarray}
win &=& \argmin_j \|\textbf{X} - \textbf{W}_j \|   \label{eq:bmu} \\ 
H_j &=& \sigma(j,win,t) \mathrm{e}^{ -\frac{\| \textbf{X} - \textbf{W}_j\|^2}{\sigma^2}} \label{eq:hidden}
\end{eqnarray}

\begin{eqnarray}
\sigma(j,win,t) &=& exp(-dist(win,i)/S(t))  \label{eq:neighbor} \label{eq:temperature} \\
S(t) &=& \sigma_{\infty} + \frac{1}{2} (\sigma_0 - \sigma_{\infty}) (1+cos \frac{\pi t}{t_{\infty}}) \nonumber
\end{eqnarray}

For a high dimensional input $\textbf{X} \in \mathbb{R}^d$, STA selects the best matching unit, $win$  among all the reference vectors associated with the hidden units of STA as in Eq. \ref{eq:bmu}, where $\textbf{W}_j \in \mathbb{R}^d$ is the reference vector associated with the $j$-th hidden unit. In Eq.\ref{eq:temperature} $\sigma_{0} > \sigma_{\infty} > 0$ are the initial and final values of the annealing term, $t$ is the current epoch, while $t_{\infty}$ is the termination epoch.

The values of the $k$-th decoder neuron, $O^{dec}_k $, and the $l$ label neuron, $0^{cls}_l$  in the output layer are defined in Eq. \ref{eq:out} where $f(x)=\frac{1}{1+e^{-x}}$

\begin{eqnarray}
O^{dec}_k &=& f(({\textbf{V}^{enc}_k})^t \textbf{H}) \nonumber  \\
O^{cls}_l &=& f((\textbf{V}^{cls}_l)^t \textbf{H})  
\label{eq:out}
\end{eqnarray}

Here, $\textbf{V}^{enc}_k$ denotes the weight vector leading from the hidden layer to the $k$-th decoder neuron,  $\textbf{V}^{cls}_l$ denotes the weight vector leading from the hidden layer to the $l$-th class neuron in the output layer, while $\textbf{H}=(H_1, H_2, \cdots, H_{N_{hid}})^t$ is the hidden layer output vector in which $N_{hid}$ is the number of hidden neurons.

The cost function is defined in Eq.  $\ref{eq:cost}$, in which $0 \leq \kappa \leq 1$ is the mixing coefficient, Here,  $\kappa=0$  generates an autoencoder, while $\kappa=1$ generates a classifier. 

\begin{equation}
L = \frac{(1-\kappa)}{2} \sum_{k}  ({O}^{dec}_k - X_k )^2 + \frac{\kappa}{2} \sum_{l} (O^{cls}_l - T_l )^2 \label{eq:cost}
\end{equation}

Applying stochastic gradient descent, the modifications of connection weights from the hidden layer to the output layers are calculated from the gradients as follows.

\begin{eqnarray}
\Delta \textbf{V}^{dec}_k &=& \frac{\partial L}{\partial \textbf{V}^{dec}_k } \nonumber \\
                                   &=&  (1-\kappa) (O^{dec}_k - X_k) O^{dec}_k (1-O^{dec}_k) \textbf{H} \nonumber \\
                                   &=&   (1-\kappa) \delta^{dec}_k \textbf{H}  \\ \label{eq:modifydec} 
\Delta \textbf{V}^{cls}_l &=& \frac{\partial L}{\partial \textbf{V}^{cls}_l }  \nonumber \\
                                   &=&  \kappa (O^{cls}_l - T_l) O^{cls}_l (1-O^{cls}_l) \textbf{H} \nonumber \\
                                   &=&  \kappa \delta^{cls}_l \textbf{H}  \label{eq:modifydec}  \label{eq:modifycls} 
\end{eqnarray}

In Eq. \ref{eq:modifydec} and Eq. \ref{eq:modifycls}, $\delta^{dec}_k$ and $\delta^{cls}_l$ are the error signals backpropagated from the $k$-th decoder neuron and the $l$-th label neuron, respectively.

The modifications of reference vectors associated with the $j$-th hidden neuron can be calculated from the gradient as follows.

\begin{eqnarray}
\frac{\partial L}{\partial \textbf{W}_j} &=& \frac{\partial L}{\partial O^{dec}_k}\frac{\partial{ O^{dec}_k}}{\partial \textbf{W}_j} + \frac{\partial L}{\partial O^{cls}_k}\frac{\partial{ O^{cls}_k}}{\partial \textbf{W}_j}\nonumber \\
                 &=& \delta^{hid}_j H_j (\textbf{X} -  \textbf{W}_j) \label{eq:modifyhid} \\
\delta^{hid}_j   &=& \frac{1}{\sigma^2} \{ (1-\kappa) \sum_k \delta^{dec}_k v^{dec}_{jk} + \kappa \sum_l \delta^{class}_l v^{cls}_{jl} \} \label{eq:deltahid}
\end{eqnarray}

In Eq.\ref{eq:deltahid}, $\delta^{hid}_j$ is error signal backpropagated to the $j$-th hidden layer.

The reference vector modification in Eq. \ref{eq:modifyhid} is similar to that of SOM, in that the difference between the input and the reference vector drives the modification and as $H_j$  includes the neighborhood function, the proximity of the hidden neuron to the best matching unit, $win$ ensures the formation of the topological structure. However, in SOM the modification is always directed toward the input $X$, while in STA the direction is controlled by the sign of $\delta^{hid}_j$, where in case of a positive $\delta^{hid}_j$the modification is identical to SOM's while a negative $\delta^{hid}_j$ repulses the reference vector away from the input vector. As $\delta^{hid}_j$ is the error signal backpropagated from the output layer, the two dimensional hidden layer in STA is self-organized based not only on the topological structure of the inputs but also their contexts that have to be generated in the output layer. It is obvious that for same inputs, different teacher signals or cost functions will generate different topological representations in the hidden layer. Hence, unlike SOM, the STA generates maps that visualizes the topological structure of the inputs in their given context.

\section{Experiments}

\begin{figure*}
\subfloat[3 Clusters]{\includegraphics[width=6 cm, height=6cm]{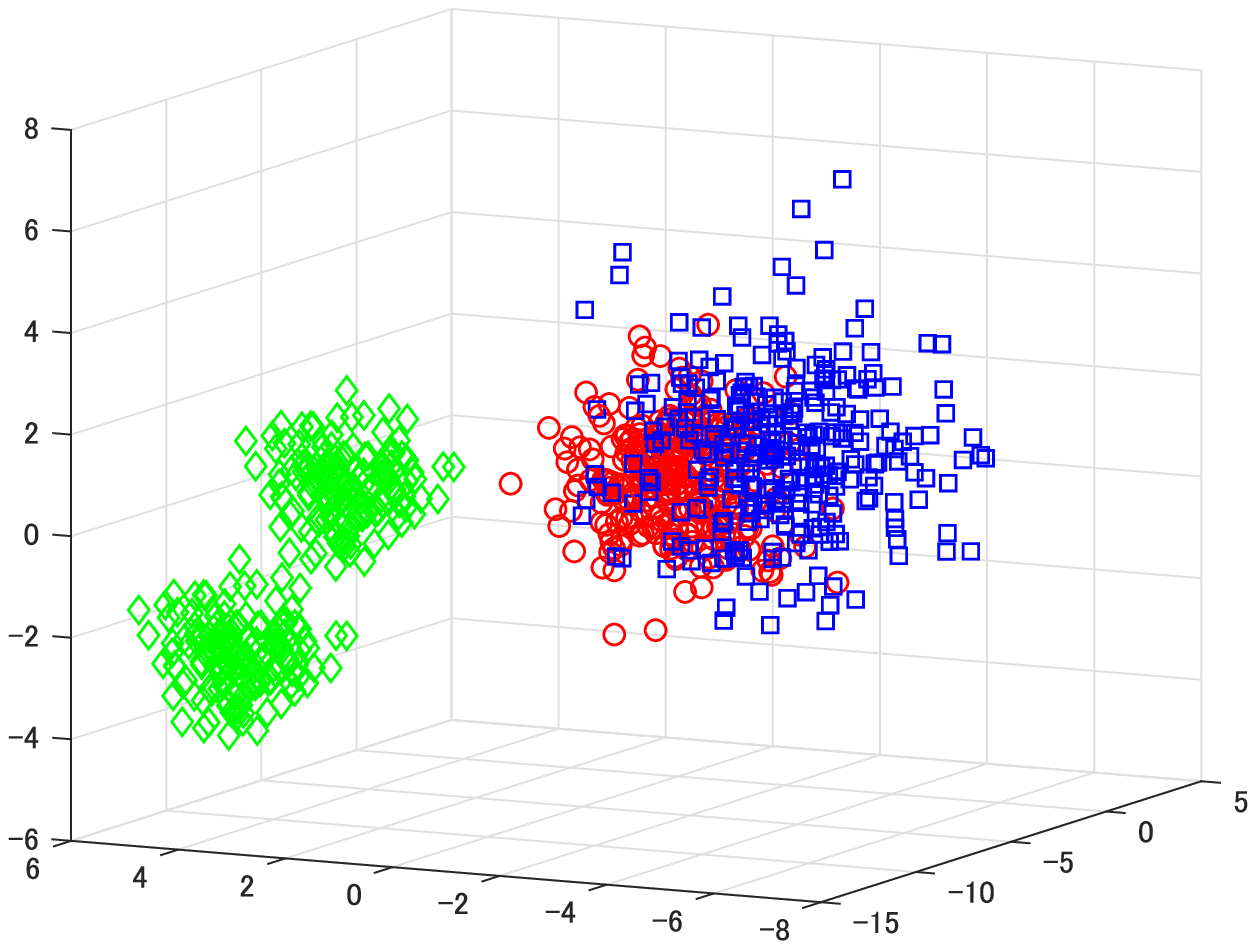}\label{fig:problem3d1}}
\subfloat[3 Clusters  ($\kappa=0$) ]{\includegraphics[width=6 cm, height=6cm]{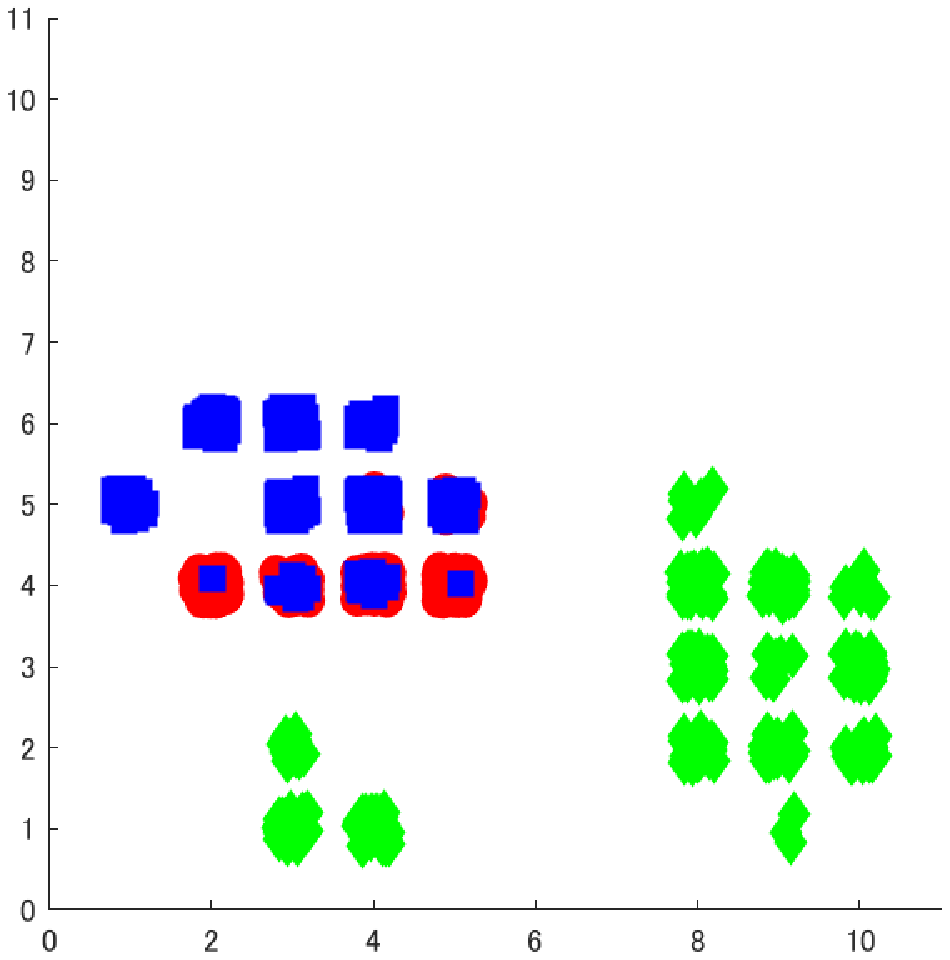}\label{fig:cluster3d0}}
\subfloat[3 Clusters  ($\kappa=1$)]{\includegraphics[width=6 cm, height=6cm]{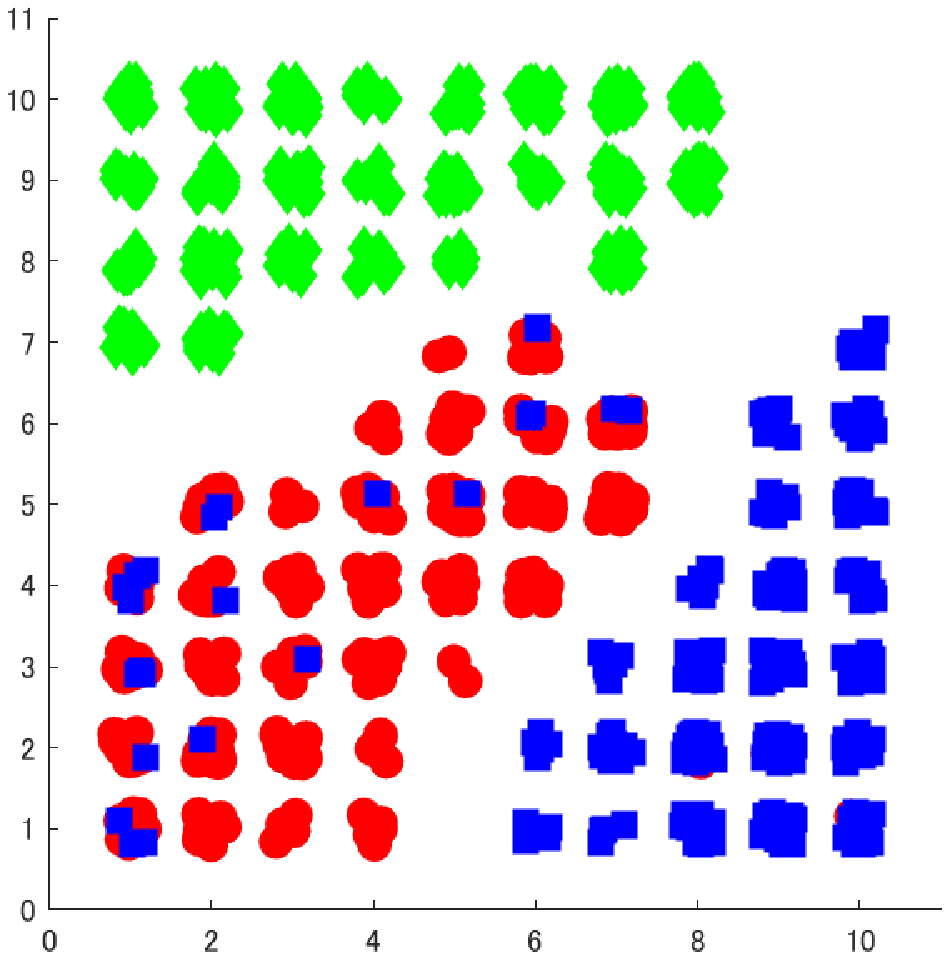}\label{fig:cluster3d1}}
\end{figure*}

In the preliminary experiment, the STA is tested against 3 dimensional toy problem shown in Fig. \ref{fig:problem3d1} where four normally distributed clusters are assigned to three classes denoted by three different colors and markers,  \textcolor{red}{\Large{$\bullet$}}, \textcolor{blue}{$\blacksquare$}  and \textcolor{green}{$\blacklozenge$}, respectively. There are many overlapping points in the two clusters represented with \textcolor{red}{\Large{$\bullet$}}s and \textcolor{blue}{$\blacksquare$}s while the rest of the two clusters are identically labeled. Figure \ref{fig:cluster3d0} visualizes the hidden representation of STA, when it was trained as an autoencoder ($\kappa = 0$) while Fig. \ref{fig:cluster3d1} visualizes the hidden representation of STA when it is trained as a classifier ($\kappa = 1$). It can be seen from these two figures that the the contexts of the data plays important roles in generating different topological representation. The autoencoder generates topological representation where adjacent clusters in their original high dimensional space are also assigned in a close proximity, while two separated clusters are assigned remotely on the low dimensional representation space. For the classifier, the labels play important role in organizing the hidden representation. It is obvious that the two originally overlapping clusters are separated, except for a few similar points.

\begin{figure*}
\subfloat[4 Clusters]{\includegraphics[width=6 cm, height=6cm]{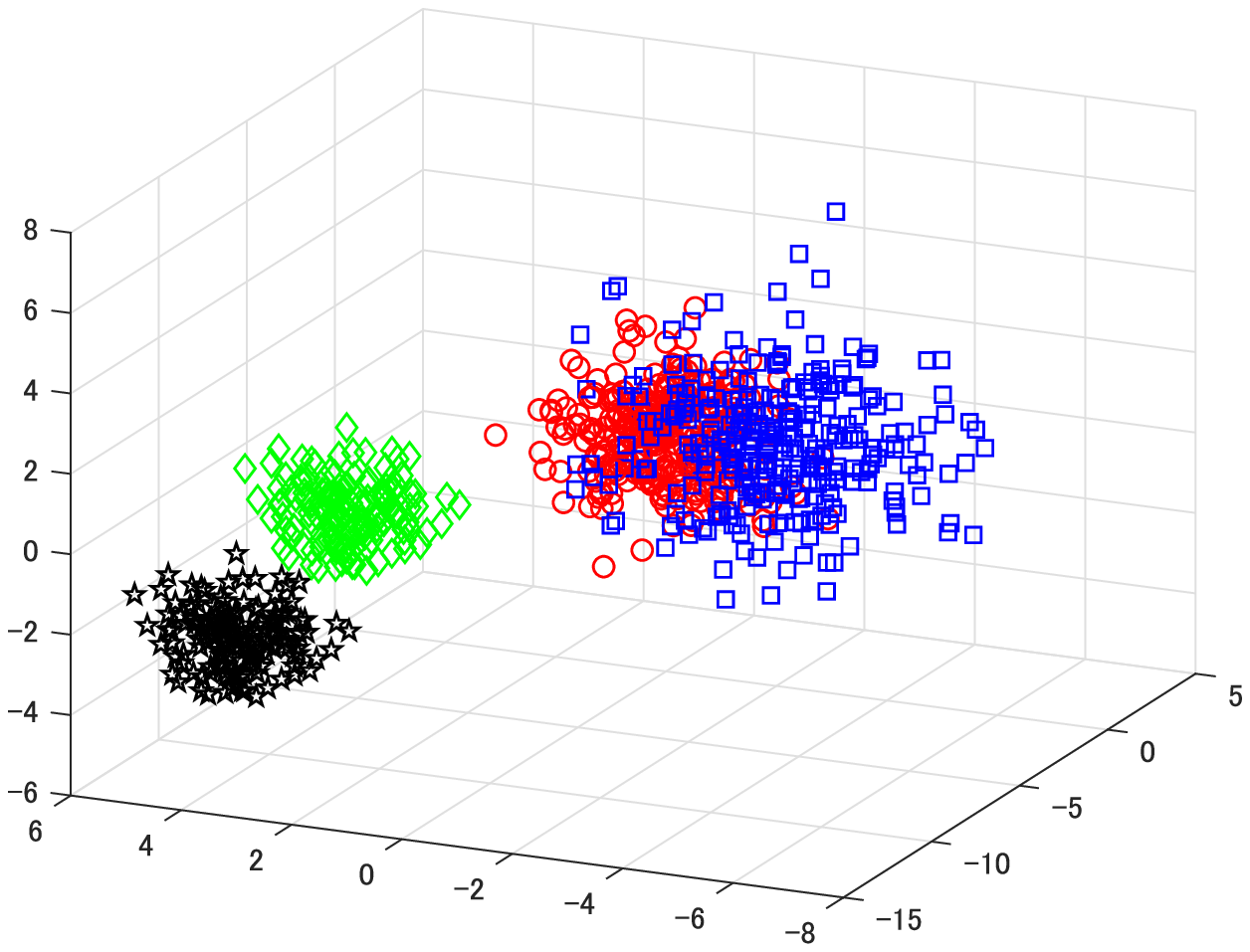}\label{fig:problem4d}}
\subfloat[4 Clusters  ($\kappa=0$) ]{\includegraphics[width=6 cm, height=6cm]{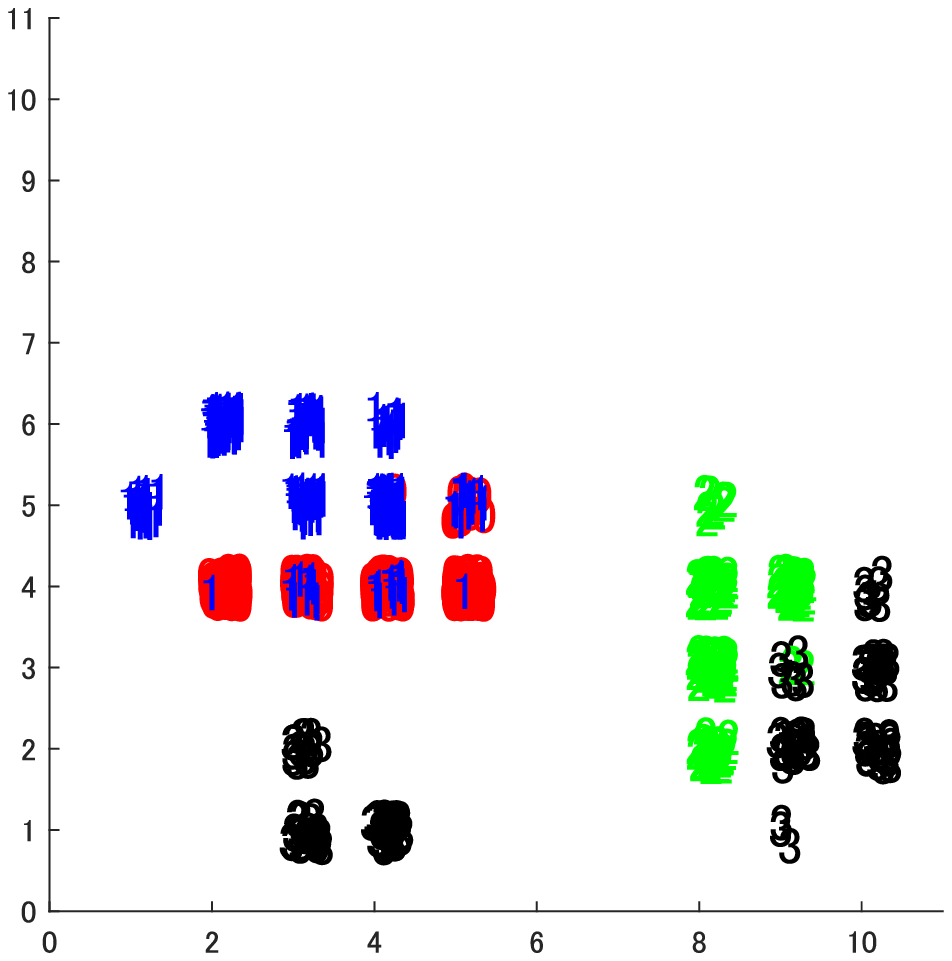}\label{fig:cluster4d0}}
\subfloat[4 Clusters  ($\kappa=1$)]{\includegraphics[width=6 cm, height=6cm]{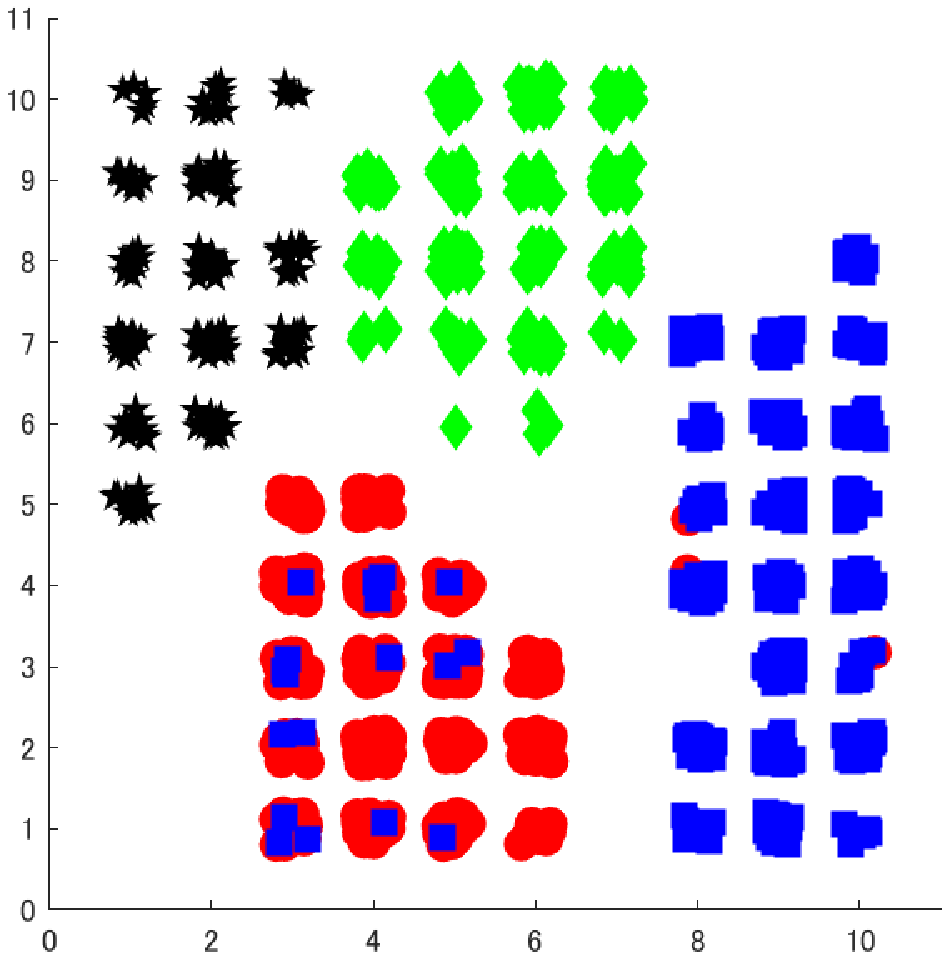}\label{fig:cluster4d1}}
\end{figure*}

The next toy problem is shown in Fig. \ref{fig:problem4d}. The data distribution is identical to the previous problem but the identically labeled two-clusters in the previous problem are labeled differently, marked as \textcolor{green}{$\blacklozenge$} and \textcolor{black}{$\bigstar$}. Figure \ref{fig:cluster4d0} visualizes the hidden representation of STA as an autoencoder. Naturally, the structure of this representation is exactly the same as the one in Fig. \ref{fig:cluster3d0}, as the labels of the data do not have any role in the learning process. However, when STA is trained as a classifier, the context of the data changes from the previous problem and thus consequently alters the distribution of the hidden representation. Now the large cluster representing the two different normal distributions with the same labels are replaced by two adjacent clusters with different labels.

The preliminary experiments indicate that the low dimensional representations of STA are influenced by the context of the data, hence now we can visualize not only the topological structure of high dimensional data as in SOM but their topological structure under a given context as well.

\begin{figure*}
\subfloat[Iris ($\kappa=0$)]{\includegraphics[width=4 cm, height=4cm]{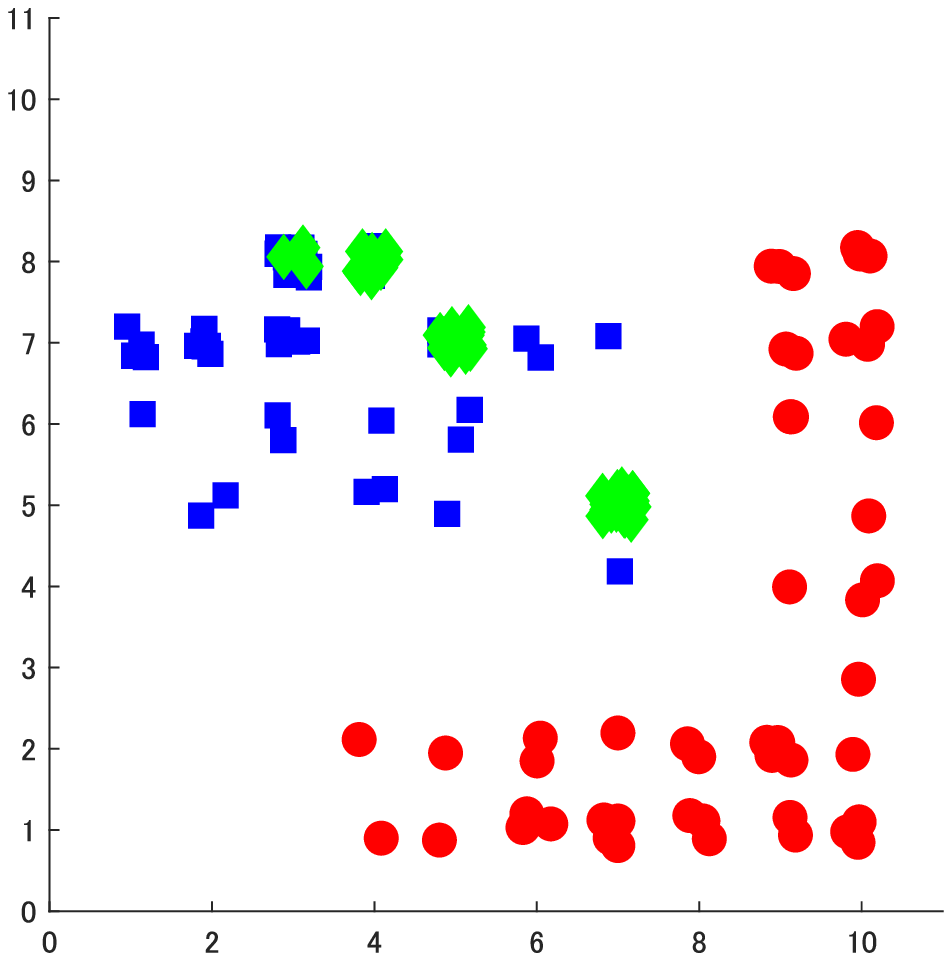}\label{fig:iris0}}
\subfloat[Iris ($\kappa=0.1$) ]{\includegraphics[width=4 cm, height=4cm]{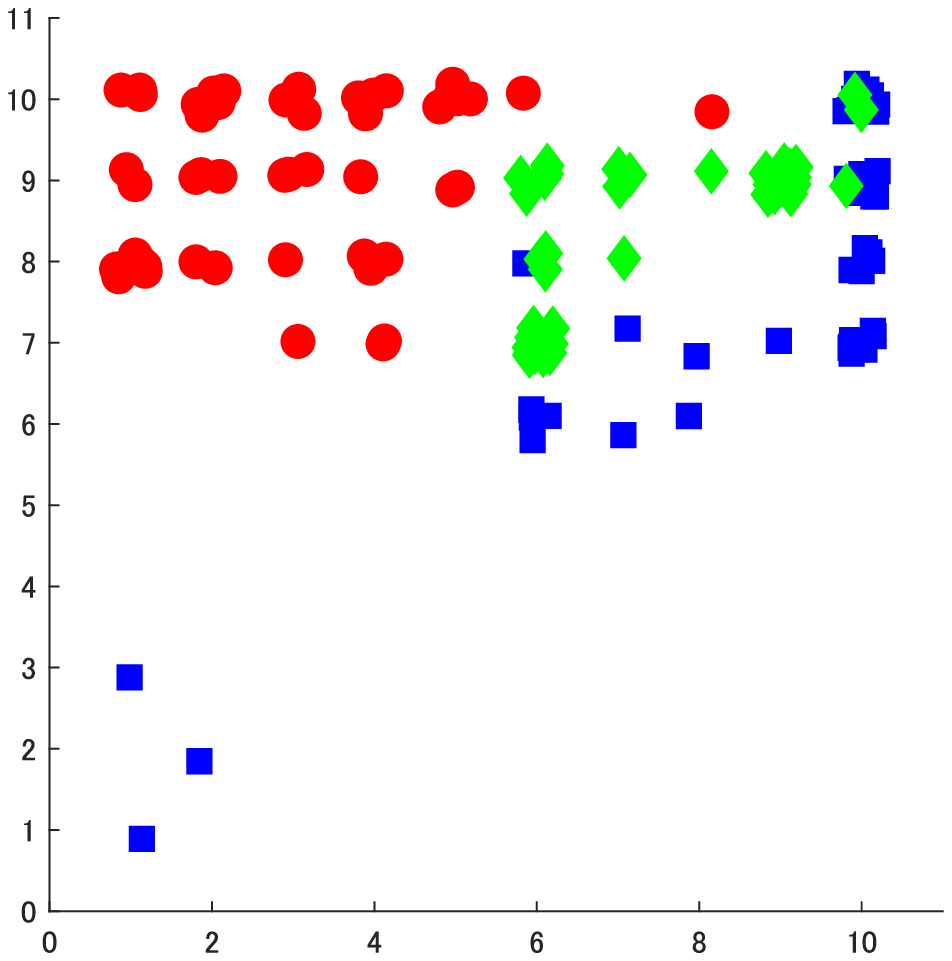}\label{fig:iris01}}
\subfloat[Iris ($\kappa=0.8$) ]{\includegraphics[width=4 cm, height=4cm]{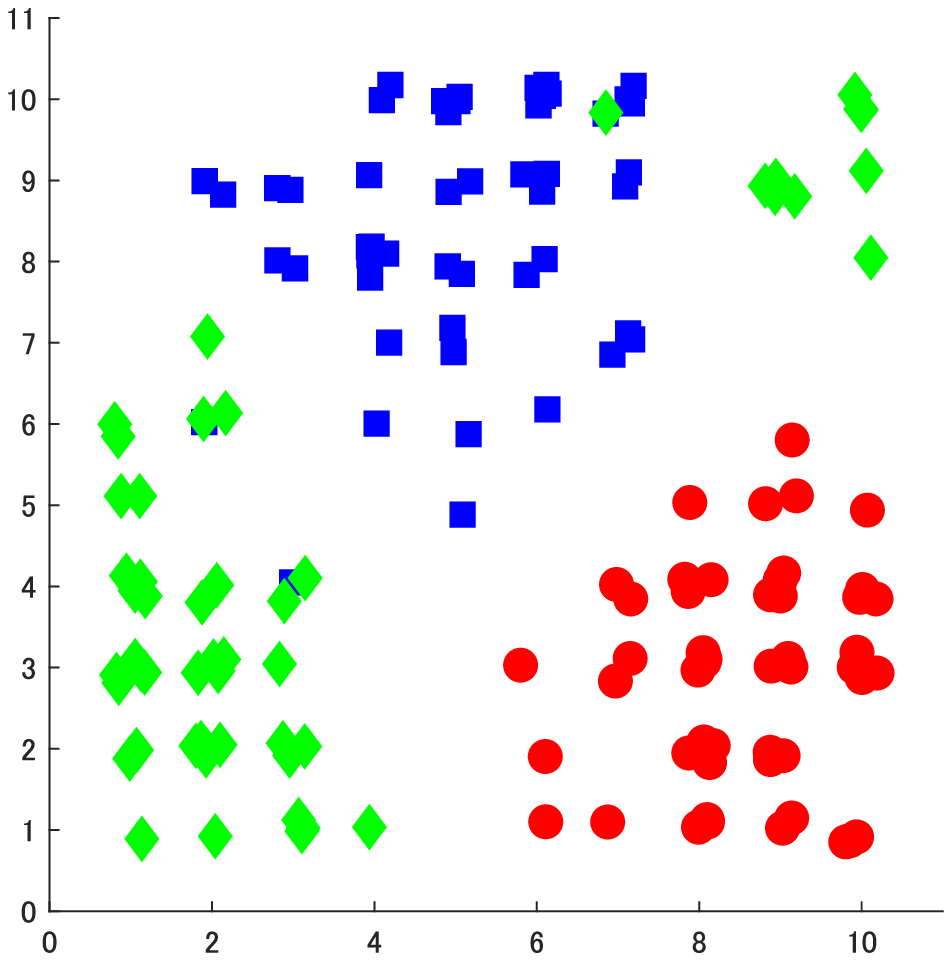}\label{fig:iris08}}
\subfloat[Iris ($\kappa=1$)]{\includegraphics[width=4 cm, height=4cm]{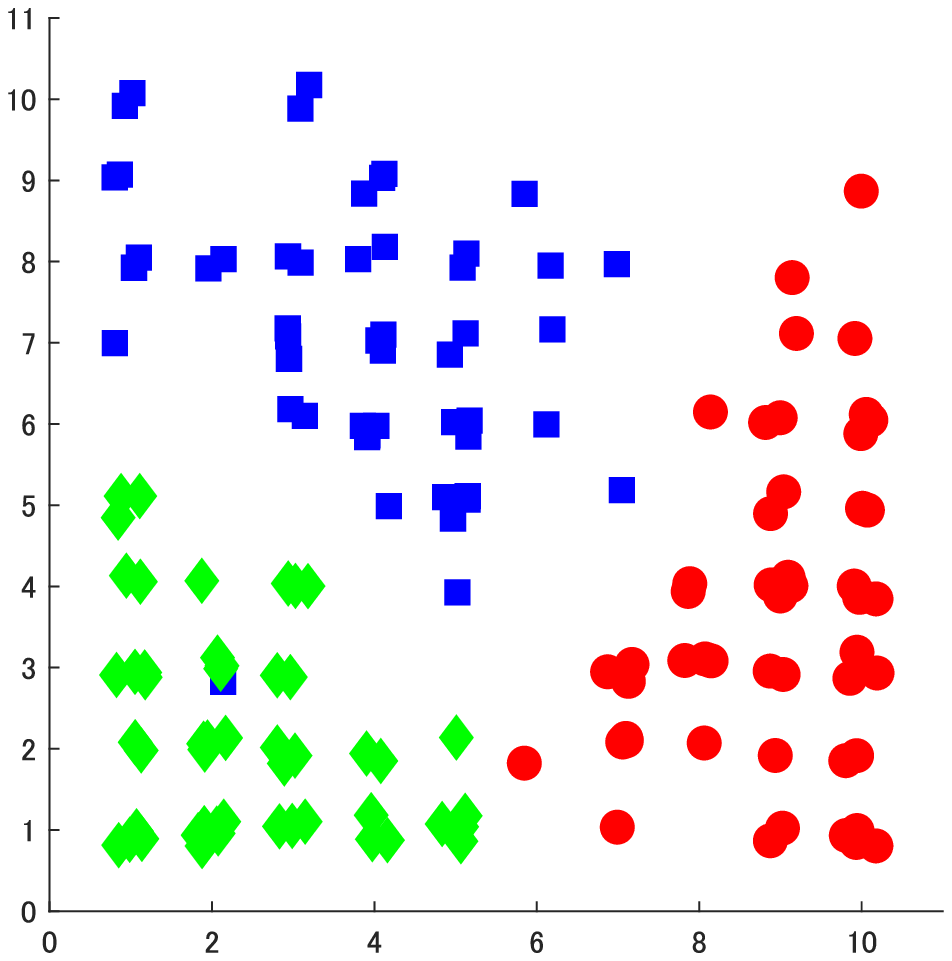}\label{fig:iris1}}
\caption{Iris Data}
\end{figure*}

To better illustrate the visualization characteristics of STA, it was trained against the well known Iris Data. These data are four dimensional and comprise of three labels where it is known that one of the classes, marked with  \textcolor{red}{\Large{$\bullet$}} is linearly separable from the other two, while those two, marked with \textcolor{blue}{$\blacksquare$} and \textcolor{green}{$\blacklozenge$}, are not linearly separable. Figure \ref{fig:iris0} shows the visualization of the hidden layer of STA trained as an autoencoder ($\kappa = 0$), where the unlabeled structure of the data is displayed. Here, the inherent characteristics of the data can be clearly observed, in which one of the classes forms a distinctive cluster separated from the other two, but the those two shares some overlapping instances.
 Increasing the mixing parameter $\kappa=0.1$ results in Fig. \ref{fig:iris01}. Here, as the labels are slightly infused into the learning process, so STA attempts to separate the two originally overlapping classes. The class-separation appears more obviously when the mixing coefficient increases, as shown in the case when $\kappa=0.8$ in Fig. \ref{fig:iris08}. When STA is trained as a classifier ($\kappa=1$) the three classes are completely separated except for a few overlapping points. 

\begin{figure*}[htbp]
\subfloat[Wine ($\kappa=0$)]{\includegraphics[width=4 cm, height=4cm]{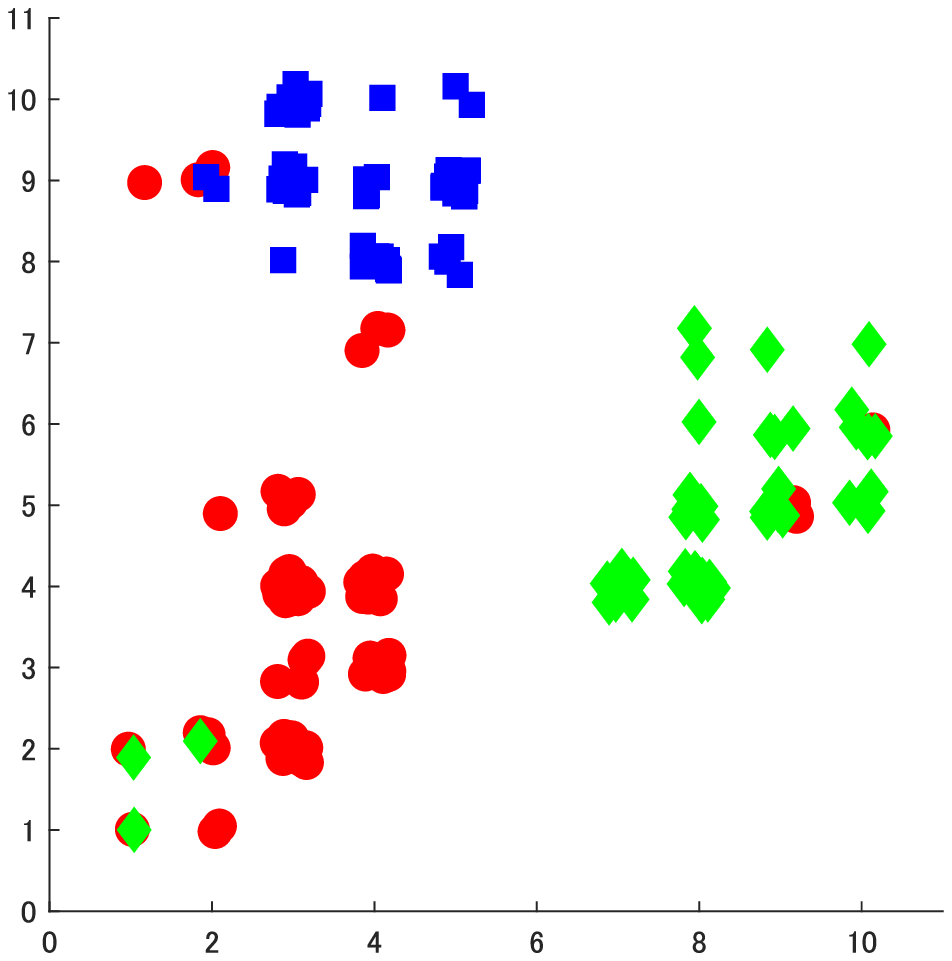}\label{fig:wine0}}
\subfloat[Wine ($\kappa=0.1$) ]{\includegraphics[width=4 cm, height=4cm]{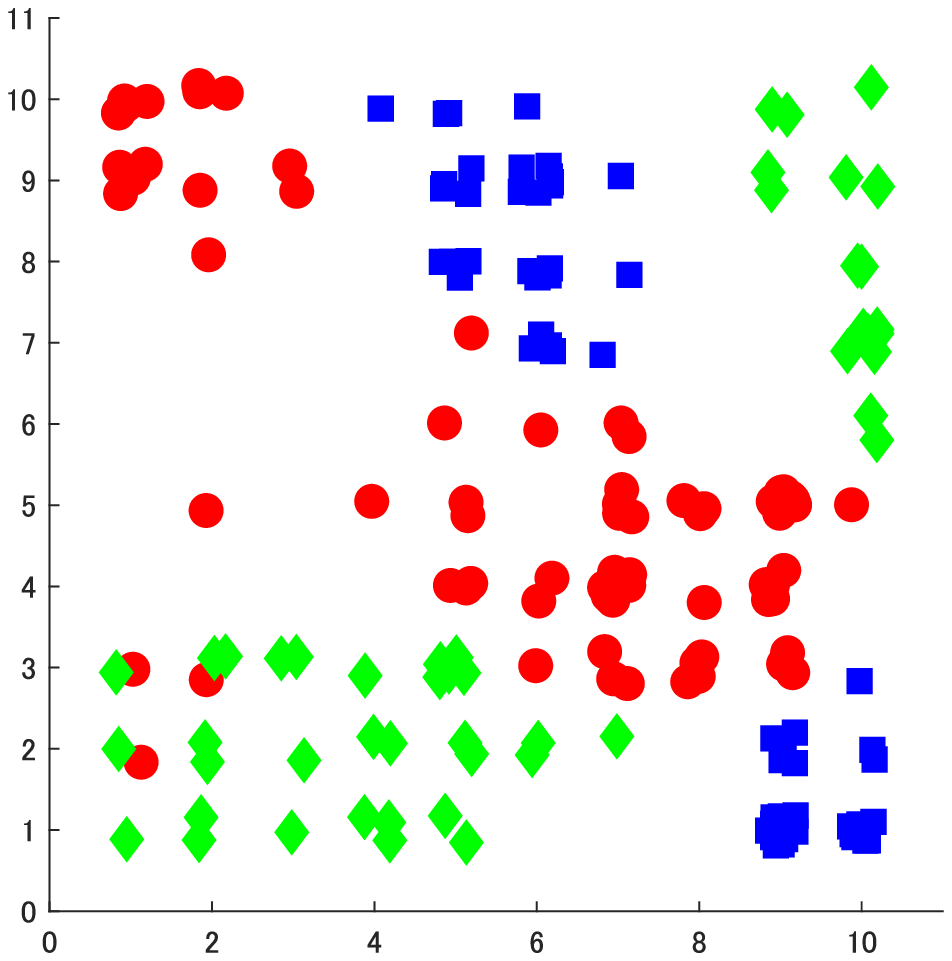}\label{fig:wine01}}
\subfloat[Wine ($\kappa=0.8$) ]{\includegraphics[width=4 cm, height=4cm]{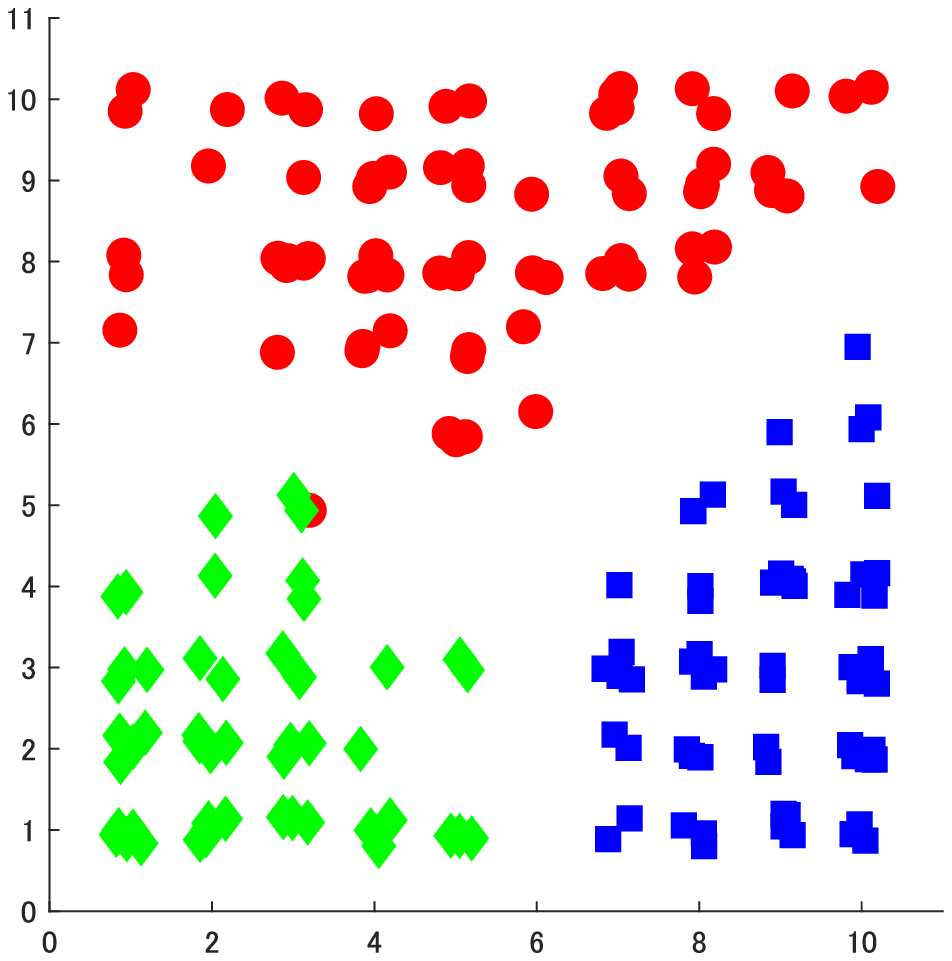}\label{fig:wine08}}
\subfloat[Wine ($\kappa=1$)]{\includegraphics[width=4 cm, height=4cm]{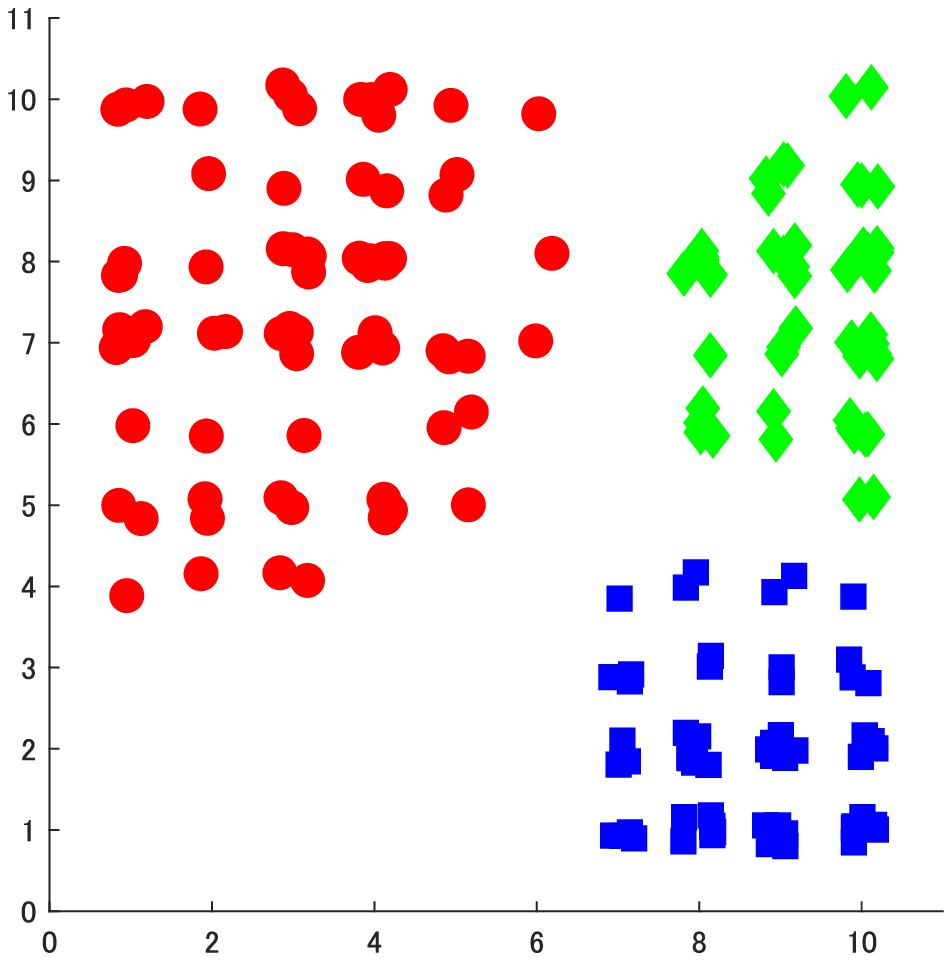}\label{fig:wine1}}
\caption{Wine Data}
\end{figure*}

Figures \ref{fig:wine0} - \ref{fig:wine1} show the visualizations of Wine Data under different learning contexts. It is clear that with the increase of $\kappa$ the hidden representations of STA shift from the formation of inherent topological structure of the data into contextual representation of the data. 

From the experiments, it can be observed how the labels of the data influence the internal organization of the neural network. Here, while an autoencoder generates topological representations capturing the original inherent structure of the data, the a classifier generates topological representations that are instrumental in predicting the labels of the data. 

\begin{figure*}
\subfloat[MNIST ($\kappa=0$) ]{\includegraphics[width=9.5cm, height=9.5cm]{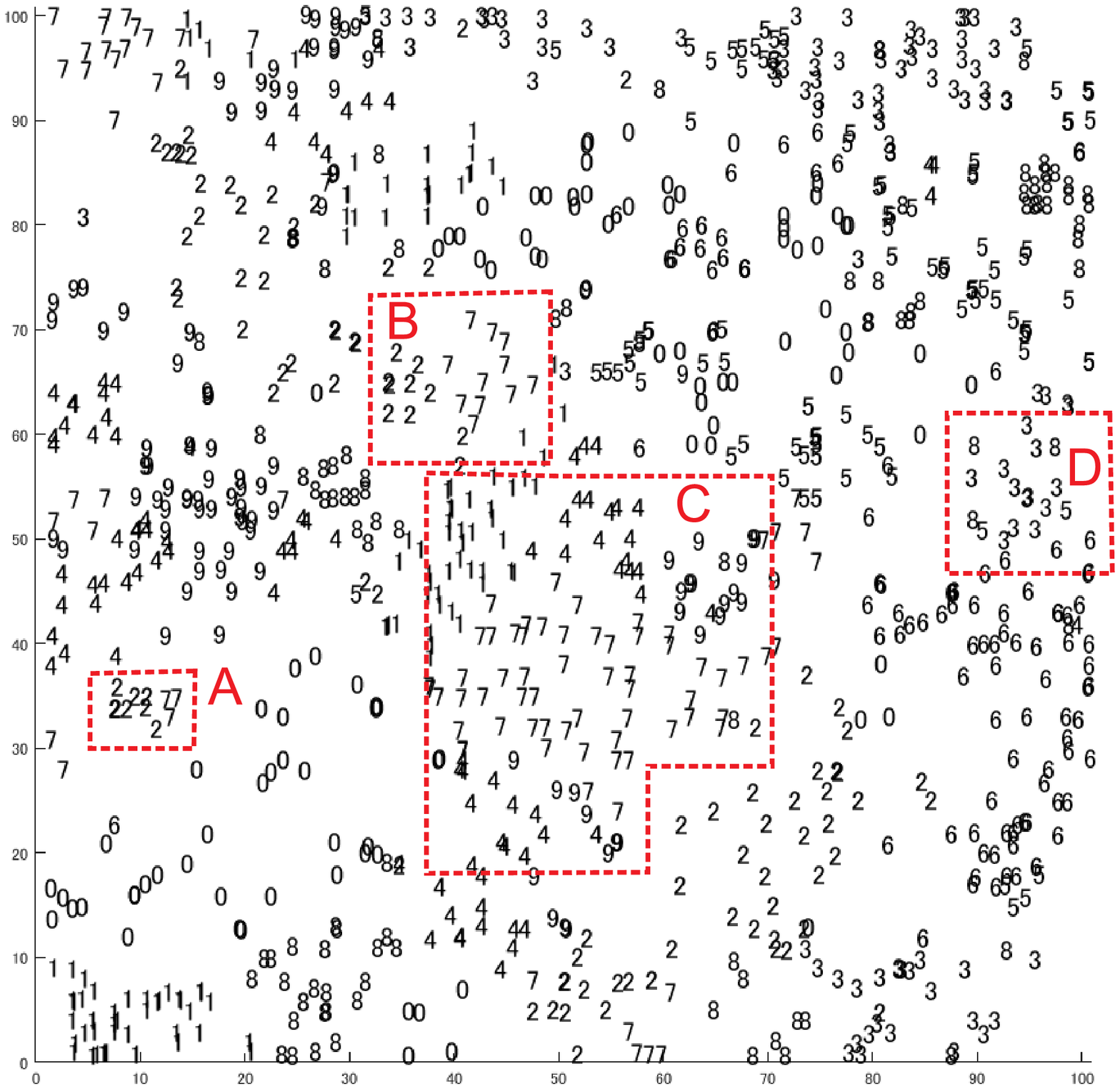}\label{fig:mnistfull0bw}}
\subfloat[MNIST ($\kappa=1$) ]{\includegraphics[width=9.5cm, height=9.5cm]{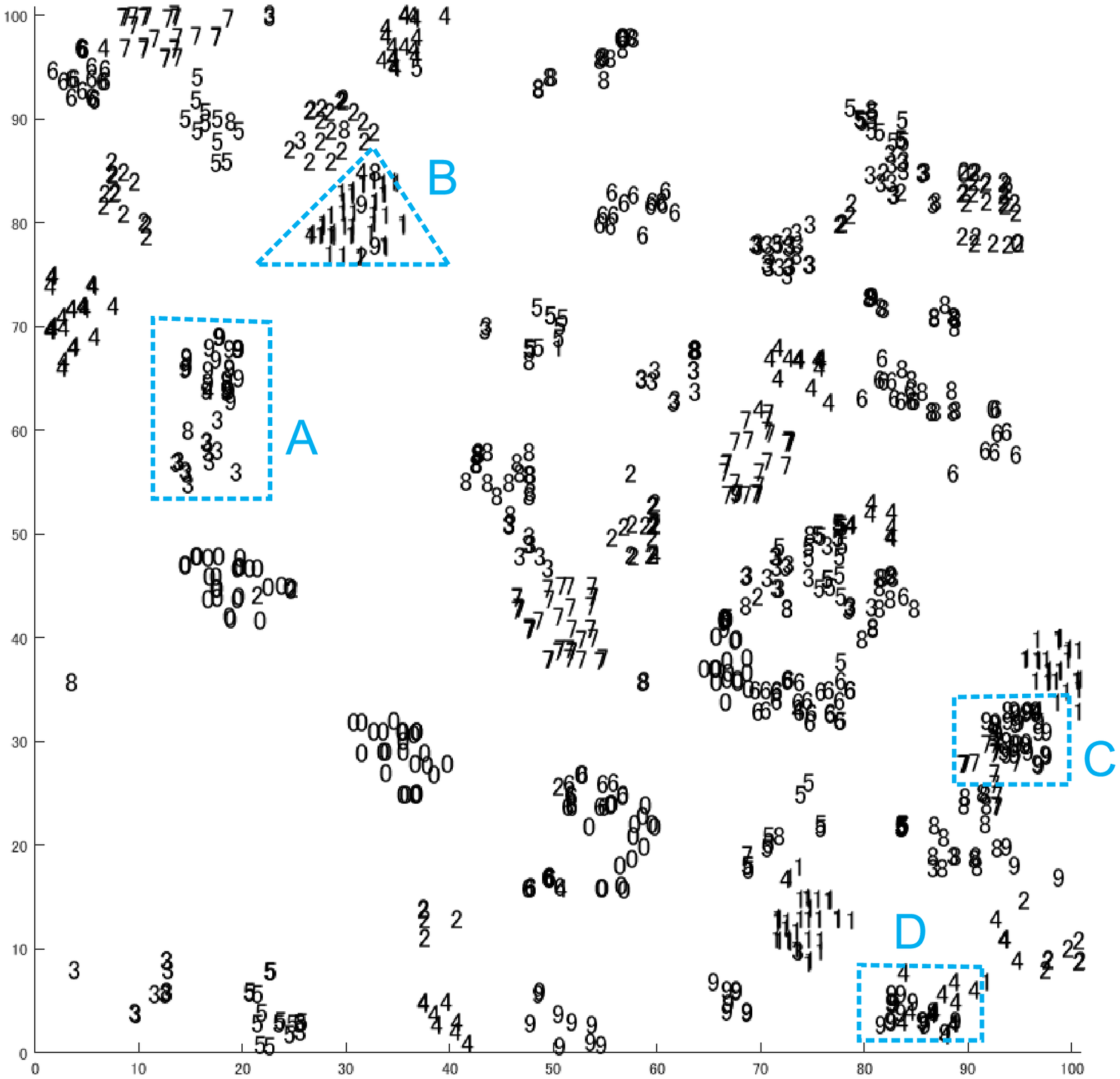}\label{fig:mnistfull1bw}}
\caption{MNIST}
\end{figure*}

In the final experiment, the STA is tested against MNIST problem. Figure \ref{fig:mnistfull0bw} shows the autoencoder's topological representation that shows the inherent structure of the handwritten digits without their label-contexts. It is natural that some different digits are similarly written, and these similarities are reflected on the topological representations of STA. For example, in group A and group B, many 2s and 7s form some clusters, while in group C, 1s, 4s, 7s and 9s are closely grouped, while in group D there are many 3s, 6s and 8s. Overall this figure shows the natural distribution of human digit writing.  The representation of STA trained as a classifier is shown in Fig. \ref{fig:mnistfull1bw}. In this figure it is obvious that the introduction of labels in the training process changes the topological representation, in that the digits are more distinctively clustered.  However, there are still some mixed clusters, for example group A contains some 3s, 8s and 9s, group B contains 1s and 9s, group C contains 7s and 9s, while group D contains 4s and 9s. The grouping of different digits into a same cluster is due to the similarity in writing different digits. The visualization does not only display the contextual distribution of the data but also offer intuitive information on input areas where the classifier is likely to perform well and other areas that are challenging for the classifier.

\section{Conclusion}
In this study, a neural network that is able to form a two dimensional topological map based not only on the high dimensional structure of the data but also their contexts is proposed. The ability to visualize high dimensional data under different contexts add flexibility in discovering obscure characteristics of the high dimensional data. As opposed to many dimensional reduction and visualization methods that are either supervised or unsupervised, STA can be flexibly trained under different context. In this paper, the STA was trained as autoencoders, where the inherent label-free characteristics of the data are captured, as classifiers, where the topological characteristics under the contextual relation of the labels are captured, or mixing both autoencoder and classifier. When the STA is trained as an autoencoder, the hidden representations encodes the natural topological structure of the data that is required to reconstruct high dimensional input in the output layer. When the STA is trained as a classifier, the hidden representation encodes a contextual topological structure that is needed to predict the labels of the high dimensional input. By controlling the mixing coefficient, an intermediate representation is formed. Observing the hidden representations under different training contexts, some insights about how the infusion of contexts changes the internal representation. For example the degree of difficulty in training a classifier can be intuitively understood. High dimensional data that have natural class-division in their autoencoder representation are likely to be an easy problem for classifier, while the difficulty of the classifier can be intuitively observed from the overlapping areas containing  contrasting samples.
In this paper, the framework for context-flexible visualization and their basic experiments have been presented. For the future works, STA is to be utilized for multi context data visualization analysis. For example in educational setting, where learning characteristics of students can be interpreted in different contexts, and be utilized to further support their learning activities. As an aspect of explainable AI, a method for explaining the topological map in a human friendly form will also be developed.

\ifCLASSOPTIONcaptionsoff
  \newpage
\fi

\bibliographystyle{IEEEtran}
\bibliography{hartonobib}






\end{document}